\documentclass[letterpaper, 10 pt, conference]{ieeeconf}  

\IEEEoverridecommandlockouts                              

\usepackage{color}
\usepackage{epsfig}
\usepackage{graphicx}
\usepackage{algorithm,algorithmic}

\usepackage{adjustbox}
\usepackage{array}
\usepackage{booktabs}
\usepackage{colortbl}
\usepackage{float,wrapfig}
\usepackage{framed}
\usepackage{hhline}
\usepackage{multirow}
\usepackage{makecell}
\usepackage[percent]{overpic}

\usepackage{amsmath,amsfonts,amssymb}
\usepackage{amsthm} 
\usepackage{bm}
\usepackage{nicefrac}
\usepackage{microtype}
\usepackage{contour}
\usepackage{courier}
\usepackage{siunitx}

\usepackage{changepage}
\usepackage{extramarks}
\usepackage{fancyhdr}
\usepackage{lastpage}
\usepackage{setspace}
\usepackage{soul}
\usepackage{xspace}
\usepackage{cuted}
\usepackage{fancybox}
\usepackage{afterpage}
\usepackage{gensymb}

\usepackage[breaklinks=true,colorlinks,backref=True]{hyperref}
\hypersetup{colorlinks,linkcolor={black},citecolor={MSBlue},urlcolor={magenta}}
\usepackage{url}
\usepackage{quoting}
\usepackage{epigraph}

\usepackage{enumerate}
\usepackage{paralist,tabularx}
\usepackage{comment}
\usepackage{pdfpages}
\usepackage{caption}  

\usepackage{pifont}

\usepackage{MnSymbol}

\usepackage[para]{footmisc}
\usepackage{cite}

\makeatletter
\DeclareRobustCommand\onedot{\futurelet\@let@token\@onedot}
\def\@onedot{\ifx\@let@token.\else.\null\fi\xspace}

\def\etal{et al\onedot}

\makeatother

\definecolor{MyDarkBlue}{rgb}{0,0.08,1}
\definecolor{MyDarkGreen}{rgb}{0.02,0.6,0.02}
\definecolor{MyDarkRed}{rgb}{0.8,0.02,0.02}
\definecolor{MyDarkOrange}{rgb}{0.40,0.2,0.02}
\definecolor{MyPurple}{RGB}{111,0,255}
\definecolor{MyRed}{rgb}{1.0,0.0,0.0}
\definecolor{MyGold}{rgb}{0.75,0.6,0.12}
\definecolor{MyDarkgray}{rgb}{0.66, 0.66, 0.66}
\definecolor{MyPink}{rgb}{1, 0.75, 0.79}
\definecolor{GreenStarColor}{rgb}{0.54, 0.84, 0.41}
\definecolor{MSBlue}{rgb}{0, 0.35, 0.49}

\newcommand{\mylisttight}{\begin{list}{$\bullet$}
    {\leftmargin4mm \itemsep0pt \itemindent0mm \topsep0mm \parsep0.5mm}
}

\definecolor{ourscolor}{rgb}{0.254, 0.411, 0.882}
\definecolor{fullcolor}{rgb}{0.501, 0.501, 0.501}
\definecolor{thritycolor}{rgb}{0.0, 0.501, 0.003}
\definecolor{diffcolor}{rgb}{1.0, 0.647, 0.0}

\newcommand{\hula}{\textbf{\textcolor{red}{\texttt{HULA-offline}}}}
\newcommand{\thriftydagger}{\textbf{\textcolor{thritycolor}{\texttt{ThriftyDAgger}}}}
\newcommand{\diffdagger}{\textbf{\textcolor{diffcolor}{\texttt{Diff-DAgger}}}}
\newcommand{\ours}{\textbf{\textcolor{ourscolor}{\texttt{Our method}}}}
\newcommand{\fulltraj}{\textbf{\textcolor{fullcolor}{\texttt{Avg.\;Full-traj.\;Length}}}}

\newcommand{\mystep}[1]{{\vspace{2mm}\noindent\textbf{#1}}}

\title{\huge \bf Uncertainty Comes for Free: \\ Human-in-the-Loop Policies with Diffusion Models}

\author{
Zhanpeng He\authorrefmark{1}\authorrefmark{2}, %
Yifeng Cao\authorrefmark{1}\authorrefmark{3}, %
Matei Ciocarlie\authorrefmark{4} %
\thanks{}
}

\begin{document}

\maketitle

{
  \renewcommand{\thefootnote}%
    {\fnsymbol{footnote}}
  \footnotetext[1]{indicates joint-first authorship.}
 
}

\begin{abstract}
Human-in-the-loop robot deployment has gained significant attention in both academia and industry as a semi-autonomous paradigm that enables human operators to intervene and adjust robot behaviors at deployment time, improving success rates. However, continuous human monitoring and intervention can be highly labor-intensive and impractical when deploying a large number of robots. To address this limitation, we propose a method that allows diffusion policies to actively seek human assistance only when necessary, reducing reliance on constant human oversight. To achieve this, we leverage the generative process of diffusion policies to compute an uncertainty-based metric based on which the autonomous agent can decide to request operator assistance at deployment time, without requiring any operator interaction during training. Additionally, we show that the same method can be used for efficient data collection for fine-tuning diffusion policies in order to improve their autonomous performance. Experimental results from simulated and real-world environments demonstrate that our approach enhances policy performance during deployment for a variety of scenarios. \end{abstract}
\section{Introduction}
\label{sec:intro}

Human-in-the-Loop (HitL) operation is a paradigm where a human operator can intervene and assist a robot during deployment. This paradigm is seeing increasing adoption in cases where robots must continue to operate adequately even in corner cases not foreseen before deployment. 

In parallel, even as recent advances in policy learning have shown significant improvements in robustness at deployment time~\cite{chi2023diffusion, florence2022implicit, luo2024serl, prasad2024consistency}, current methods can still fail due to problems such as data distribution shift~\cite{li2024evaluating} or incomplete state observability~\cite{cong2021comprehensive}. 
To address this issue, HitL methods can be a natural fit for learning agents: the robot operates autonomously when possible, leveraging the ability of policy learning to execute complex motor control tasks. An expert operator can take over for corner cases, ensuring task success. However, deploying HitL can be labor-intensive and impractical if it implies constant monitoring of the robot's behavior by the human operator, or frequent interventions.

In this work, we propose a data-driven approach for generating HitL policies. We start from the basic HitL premise: the robot generally acts autonomously, but a human operator is available to provide teleoperation commands should the robot require them. Our method is designed to determine when the agent should request such expert assistance, making effective use of a limited number of such calls during deployment. We also remove the need for expert intervention during the training phase, as that would place a large burden on the operator. This means that the agent has no knowledge about the effect of assistance, except for the assumption that is effective for task completion. 
\begin{figure}
  \begin{center}
     \includegraphics[width=\linewidth]{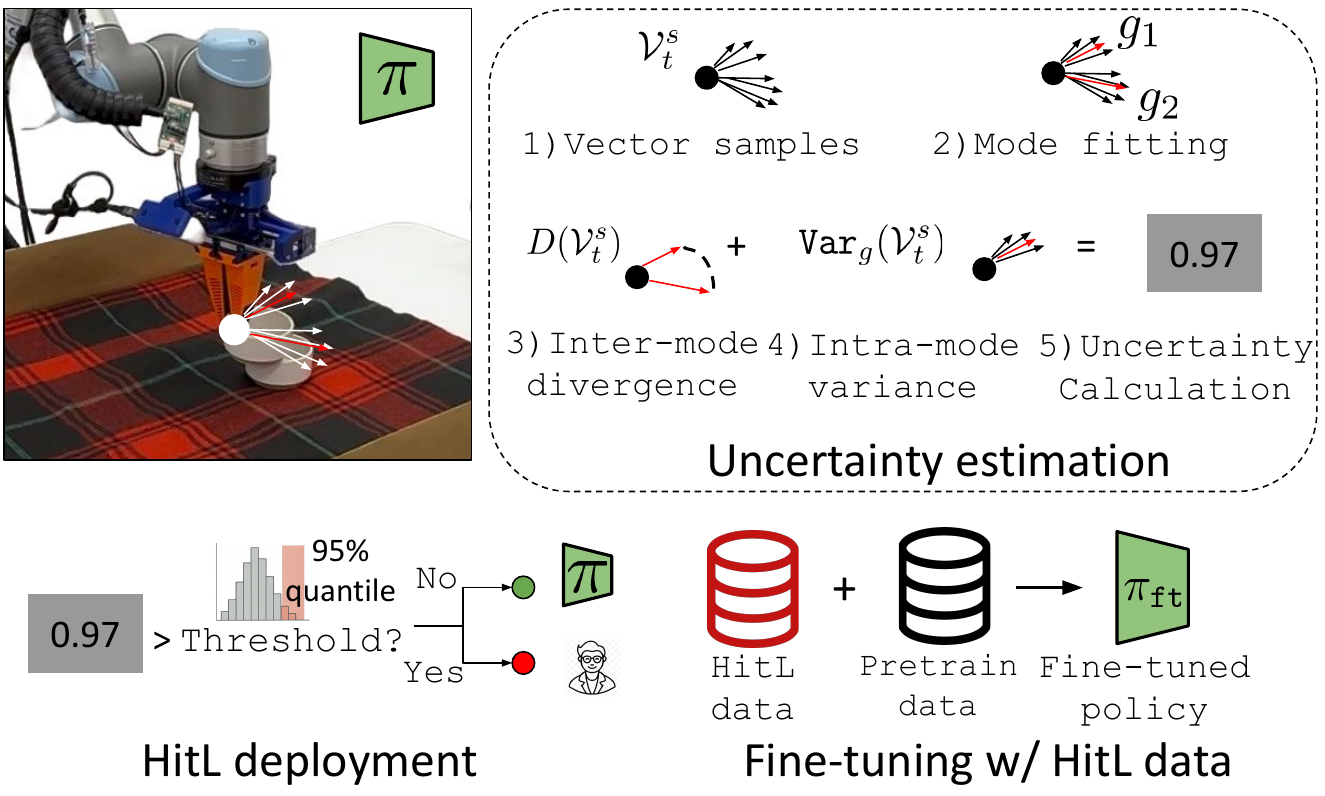}
     \vspace{-6mm}
  \end{center}
  \caption{\textbf{HitL policies with denoising uncertainty:} We propose using denoising uncertainty as a metric for deciding when to request  (human) expert assistance. Predicted de-noising vectors in end-effector position space (illustrated here via arrows on end-effector position) are collected in a vector field, whose inter-mode divergence and intra-mode variance are used to compute policy uncertainty; when this measure exceeds a threshold, operator assistance is requested. 
  We also show that ensuing teleoperation data can be used to fine-tune policies, achieving notable performance improvements with minimal additional data.\label{fig:candy}}
\vspace{-4mm}
\end{figure}

To achieve this, we utilize diffusion models~\cite{chi2023diffusion} as our policy class. Diffusion policies offer two key advantages: (1) they have demonstrated robust performance in imitation learning tasks, and (2) their generative process involves an iterative denoising mechanism which we can leverage for insight into the agent's decision-making process. 
Specifically, we use the denoising information to compute an uncertainty metric for the policy, which is then used during deployment to determine when human intervention is most beneficial (Fig.~\ref{fig:candy}). 
To achieve this, we directly leverage the noise prediction model learned during the policy training process. 
Therefore, uncertainty estimation does not require training any additional models, imposes a minimal cost at run time, and can thus be considered a ``free" byproduct of diffusion policy training. 

A key feature of our metric is that it remains consistent across different tasks: \textit{we use the same fixed quantile-based threshold, estimated from a validation set, for all deployments}, avoiding per-task calibration and underscoring the robustness and generality of our approach.
Finally, we show that data collected during the operator interventions can be incorporated back into training through a fine-tuning process that further improves policy performance. Our main contributions are as follows:
\begin{itemize}
    \item We propose a novel method for HitL policy execution using uncertainty estimation with diffusion policies. Our method does not require human-robot interaction during training, and incurs minimal computational overhead during deployment. 
    \item We validate our method across multiple types of deployment challenges, in both simulated and real environments. Experimental results show that our approach requires fewer human interventions and achieves higher task performance compared to alternative learning-based HitL agents.
    \item We also show that our uncertainty-based state identification method can be utilized to collect targeted fine-tuning data, yielding performance improvements with smaller datasets compared to collecting additional full-trajectory demonstrations.  
\end{itemize}

\section{Related Work}
\label{sec:related}

HitL approaches have been widely explored to enhance robot manipulation by incorporating various human feedback modalities, such as interventions~\cite{mandlekar2020human, spencer2020learning}, preferences~\cite{lee2021pebble}, rankings~\cite{pmlr-v97-brown19a}, scalar feedback~\cite{macglashan2017interactive}, and gaze~\cite{zhang2020human}. Recent works like HIL-SERL~\cite{liu2022robot} and Sirius~\cite{luo2024hilserl} show strong performance by leveraging human input during training. HitL methods are also common in autonomous driving, e.g., ZOOX’s re-routing system~\cite{zoox2024} allows operators to assist in challenging scenarios. However, most approaches rely on continuous human supervision and monitoring when assistance is actually needed during deployment. We propose an uncertainty-aware diffusion model that selectively requests human help in high-uncertainty states, reducing the need for constant oversight. Unlike HULA~\cite{he2024hula}, which models uncertainty for online RL, our approach derives HitL policies from offline data, enabling more efficient and scalable deployment.

Our work is closely related to interactive imitation learning (IIL), where a learning agent queries an expert for additional labels during policy execution and augments the training dataset with expert demonstrations~\cite{ross2011reduction}.
An effective data collection strategy for IIL is human-gated DAgger~\cite{kelly2019hg}, which relies on a human to continuously monitor and intervene during robot execution. However, such continuous supervision is inefficient and undermines the goal of robot learning. To make HitL systems practical, the robot must be strategic in when and how it requests human assistance.
Prior work has explored budgeting expert queries in IIL. For example, Hoque \etal reduce human effort by constraining robot queries using models of limited human attention~\cite{hoque2023fleet} or fixed intervention budgets~\cite{hoque2021thriftydagger}. 
However, these methods require extra training procedures during training or deployment, increasing computational demands. This overhead is especially problematic during policy execution, where it may introduce additional latency. In contrast, our approach avoids additional training and supports efficient parallelization during deployment, resulting in minimal runtime overhead.
Other approaches leverage action consistency~\cite{agia2024unpacking}, diffusion loss~\cite{lee2024diff}, or online conformal prediction~\cite{zhao2025conformalized} to decide when to query the human. One crucial aspect of this line of work is the selection or online-tuning of their thresholds.
To minimize human efforts in a HitL pipeline, our method leverages the multi-modal nature of human demonstrations to detect critical states where the robot is uncertain. We demonstrate that our method is robust to threshold selection in different scenarios.

\section{Method}
\label{sec:method}

We begin the description of our method with a short recap of diffusion policies, specifically the action denoising process which we will then leverage to introduce our metric for quantifying uncertainty. After that, we describe how this metric can be used by a HitL diffusion policy agent to determine when to request operator assistance, and finally how teleoperation data obtained through this method can in turn be used to fine-tune the original policy.

Diffusion policies generate actions through an action-denoising process, leveraging denoising diffusion probabilistic models (DDPM). A DDPM models a continuous data distribution $p(a^0)$ as reversing a forward noising process from $a^0$ to $a^K$, where $a^K$ is Gaussian noise sampled from $\mathcal{N}(0, \sigma^2\mathcal{I})$. The generative process $\pi(\mathrm{a_t}|\mathrm{o_t})$, where $\mathrm{o_t}$ and $\mathrm{a_t}$ are robot observations and actions at time step $t$, starts by sampling an action $a^{K}_t$ as random noise, and then iteratively denoises it using:
\begin{equation}
a^{k-1}_t = \beta(a^{k}_t - \gamma \epsilon_{\theta}(o_t, a^{k}_t, k) + \mathcal{N}(0, \sigma^2\mathcal{I})) 
\end{equation}
where $\beta$, $\sigma$ and $\gamma$ are functions of iteration step $k$. $\epsilon_{\theta}(o,a^k,k)$ is a learned model that predicts the noise to be removed at each denoising step.

\subsection{Denoising-based uncertainty metric}

To estimate the uncertainty of a diffusion-based agent, our method leverages the generative process described above. In particular, we assume that our diffusion policy is operating on task space control, which is a very common case in recent diffusion-based robot policy learning methods~\cite{ren2024diffusion, ze20243d, ke20243d}, and outputs absolute end-effector poses as part of its action vector. In this case, the noise predicted (and removed) during the generative process can be interpreted as a vector field pointing toward the distribution for intended end-effector pose the at the next step. We can thus leverage this vector field to analyze whether the diffusion-based agent is confident about its generative target.

Our goal is to estimate an uncertainty metric $\texttt{Uncertainty}(o_{t})$ where $o_t$ is the observation at time step $t$. We begin by sampling a set of end-effector poses $\mathcal{A}^{s}_t$, where each entry $a^s_t \in \mathcal{A}^s_t$ is within a distance $r$ from the current pose. When operating in task-space position control, each of these samples can be interpreted as an action vector. We can thus feed these samples through the diffusion policy noise prediction model, and collect the predicted noise vectors: let the set $\mathcal{V}^s_t$ contain all vectors $v^s_t = \epsilon_{\theta}(o_t, a^s_t, 0)$ computed for each $a^s_t \in \mathcal{A}^s_t$. This vector field encodes directions toward the action distribution that the policy aims to recover. We will use these denoising vectors to estimate uncertainty, defined as $\texttt{Uncertainty}(o_{t}) = f(\mathcal{V}^s_t)$.

The simplest method to assess uncertainty is to consider the variance of the vector field $\mathcal{V}^s_t$. However, diffusion policies are often used for their ability to capture multi-modality in the underlying demonstrations: from any given state, there might be multiple distinct action trajectories that accomplish the desired task. Thus, the denoising vector field could reflect the multi-modal nature of the demonstration data, and naive variance estimation of the vector field may fail to capture this effect. 

\begin{algorithm}
\caption{HitL Policy Deployment}
\label{alg:uncertain}
\begin{algorithmic}[1]
    \WHILE {rollout not done}
        \STATE Sample set $\mathcal{A}^{s}_t$ of poses within radius $r$ of end-effector\\
        \STATE Compute vector field $\mathcal{V}^{s}_t$ containing $v^s_t = \epsilon_{\theta}(o_t, a^s_t, 0)$ for each $a^s_t \in \mathcal{A}^s_t$  \STATE Estimate policy uncertainty based on $\mathcal{V}^{s}_t$ using Eq. (\ref{eq:uncertainty})\\
        \IF {$\texttt{Uncertainty}(o_t) \ge \texttt{threshold}$}
            \STATE Execute $m$ steps of human input actions $a_{human}$
            \STATE Save intervention data $\{o, a_{human}\}^m$ to $\mathcal{D}_{ft}$
         \STATE \hspace{-4.5mm} \textbf{else:} Execute action $a_t$ from the policy $\pi(a_t | o_t)$
        \ENDIF
    \ENDWHILE
    \IF{fine-tune}
        \WHILE {fine-tuning not done}
            \STATE Sample a batch of data from $\mathcal{D}_{ft}$ and $\mathcal{D}_{train}$
            \STATE Optimize $\pi$ with sampled data
        \ENDWHILE
    \ENDIF
\end{algorithmic}
\end{algorithm}

To address this, we use Gaussian Mixture Models (GMMs) to capture the potentially multi-modal nature of action generation. Our method, outlined in Algorithm \ref{alg:uncertain}, starts by fitting the collected denoising vectors with $N$ GMMs, each using a different number of modes. We then select the best-fit GMM for uncertainty estimation via maximum likelihood estimation:
\[
\max_{n, \theta_{\mathrm{g}}} P(\mathcal{V}^s_t; n, \theta_{\mathrm{g}}),
\]
where $n$ is the number of modes and  $\theta_{\mathrm{g}}$ contains the parameters of the GMM. With the best-fit GMM, we then estimate the agent's uncertainty. We first evaluate the divergence between each mode:
\[
D(\mathcal{V}^s_t) = \frac{1}{n(n-1)}\sum_{i, j} 1 - S_{c}(g_i, g_j)
\]
where,
\[
S_c(g_i, g_j) = \frac{g_i \cdot g_j}{||g_i|| \cdot ||g_j||}
\]
Here, $g_i$ represents the mean of the $i^{th}$ mode and $S_c$ represents cosine similarity between two vectors.
We also evaluate the GMM variance as part of the uncertainty estimation:
\begin{align}
    \texttt{Var}_g(\mathcal{V}^s_t) = \sum_{i} p(v_{i})\texttt{Var}(v_{i})
\end{align}
where \texttt{Var} represents the variance of vector data and $v_i$ represents vector samples belongs to the $i^{th}$ mode of the GMM.
Putting them together, we can estimate the overall uncertainty as:
\begin{align}
    \label{eq:uncertainty}
    \texttt{Uncertainty}(o_t) = D(\mathcal{V}^s_t) + \alpha \texttt{Var}_{g}(\mathcal{V}^s_t), 
\end{align}
where $\alpha$ is a constant.
This uncertainty estimation considers two aspects during denoising: how diverged the target distributions are, and how much entropy there is in each of the modes.

\vspace{-0.3em}
\subsection{Uncertainty-based intervention and policy fine-tuning}
\label{sec:deployment-method}
\vspace{-0.3em}

Having defined our uncertainty metric, we can use it during deployment by setting a threshold to determine whether we to request human assistance. At every state, the agent computes its own uncertainty and, if the level of uncertainty exceeds the threshold, the agent requests that the operator take control and teleoperate the system for several steps, until uncertainty returns below the threshold.

In addition, our method can also be used to collect data to further fine-tune the policy. This allows for better performance in the next policy execution. To fine-tune a policy, we save the observation and action pairs $\{\mathrm{O}, \mathrm{A}\}$ when a human operator is intervening with the robot and use this data set to fine-tune the underlying diffusion policy. To avoid catastrophic forgetting \cite{ball2023efficient}, we sample from both the fine-tuning dataset $\mathcal{D}_{ft}$ and pretraining dataset $\mathcal{D}_{train}$. For each mini-batch, we ensure $50\%$ are from $\mathcal{D}_{ft}$. Our approach implicitly means that this fine-tuning data specifically addresses the areas of state space where the agent's uncertainty is high, since that is where operator assistance is requested.

Putting all components together, the method contains three main steps: 1. train a diffusion policy; 2. deploy the policy, and request operator control if policy uncertainty estimated by our metric exceeds a preset threshold; 3. (optional) use human intervention data to fine-tune the diffusion policy.

\vspace{-0.3em}
\section{Experiments}
\label{sec:sim-exp}
\begin{figure}
    \centering
    \includegraphics[width=\linewidth]{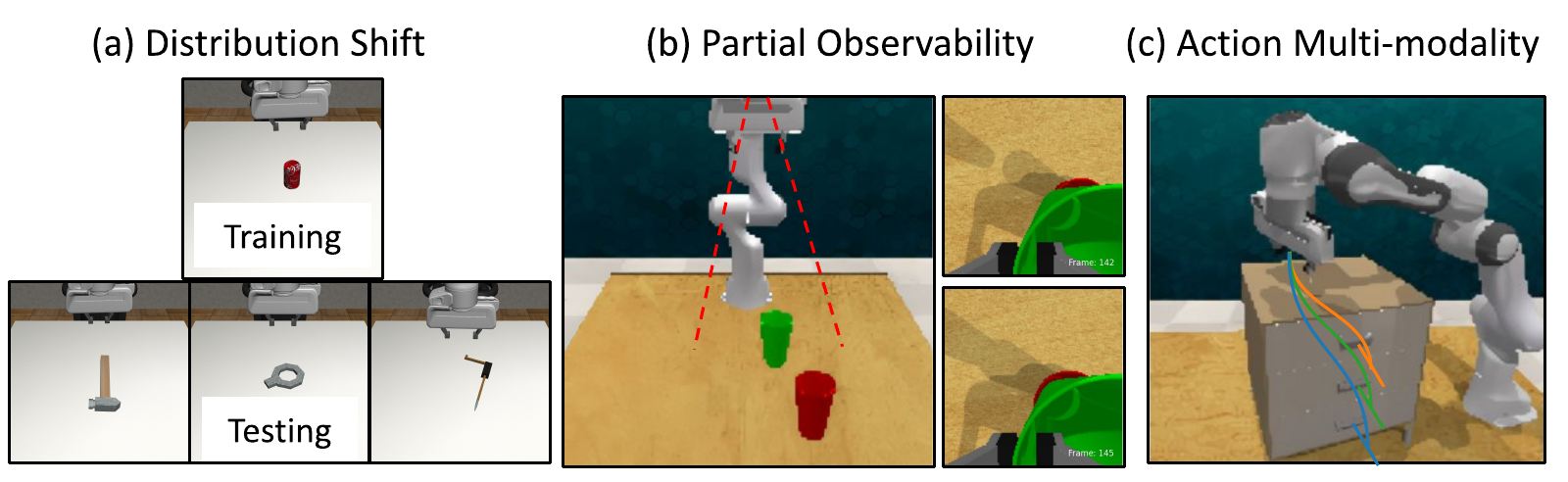}
    \caption{\textbf{Experiments in simulated environments.} We consider three scenarios during policy deployment. (a) Distribution shift; (b) Partial observability (c) Action multi-modality. \label{fig:envs}\vspace{-2em}}
\end{figure}

To test applicability of this framework, we consider three types of deployment issues that typically cause uncertainty for learning-based agents.
\textbf{Case 1: Data distribution shift}, such as visual observation distribution shift caused by change of lighting conditions, or a change in environment dynamics due to interaction with novel objects.
\textbf{Case 2: Incomplete state observability}, commonly approached by redesigning, adding or moving sensors, but difficult to tackle in the general case. 
\textbf{Case 3: Incorrect choice between different action modes}, where the agent is presented with a discrete choice between two or more action trajectory modes equally well represented during training. While diffusion policies are naturally well-equipped to make such choices, task under-specification can lead to the selection of the incorrect action mode for the given goal.

During policy execution, these problems may not be present in all states -- many states are easy to make decisions for, and require no human intervention (e.g. moving the arm in free space). The goal of our metric is to identify when the issues described above arise, and selectively request help. For Case 3 above, we posit that a few steps under teleoperator control can ``steer" the policy towards the desired mode, after which autonomous operation can resume. Case 1 lends itself well to fine-tuning based on the novel data collected during teleoperation. Finally, we expect Case 2 to be the most difficult, since correct decision making is impossible without changing the available observation. We design our experiment set to test a range of scenarios covering these situations.

\vspace{-0.5em}
\subsection{Evaluation and Baselines}
\vspace{-0.3em}
We validate our method across the three types of deployment challenges in both simulated and real environments. In our test scenarios, full teleoperation generally succeeds, with sufficient human intervention achieving near 100\% task success. However, a key goal of HitL deployment is efficiency: assistance should be requested conservatively to minimize unnecessary interruptions. 

Our evaluation thus focuses on two core aspects. First, we measure the efficiency of human-robot interaction by tracking the required frequency of human interventions to achieve 100\% task success. Second, we assess the improvement in task performance enabled by human assistance and policy fine-tuning, quantifying the impact of integrating human feedback. We compare our approach against three state-of-the-art baselines that incorporate uncertainty estimation into HitL frameworks:
\begin{compactitem}
    \item \thriftydagger~\cite{hoque2021thriftydagger}, which uses a model-ensemble-based OOD detection and a risk metric learned via Bellman updates on test-time data. 
    \item \diffdagger~\cite{lee2024diff}, which uses predicted actions from a diffusion policy to compute the diffusion loss, and, based on it, a metric to decide requesting human assistance.
    \item \hula~\cite{he2024hula}, which produces an RL-based HitL policy by explicitly estimating the variance of state values. To make it comparable to our method, where only offline datasets are available during training, we adapt it to offline RL by implementing an offline variant using Conservative Q-Learning (CQL)~\cite{kumar2020conservative}.
\end{compactitem}
We also try other metrics, namely Sentinel~\cite{agia2024unpacking} and diffusion output variance, that can be potentially adapted into a HitL pipeline. We summarize discussions about our attempts in Appendix~\ref{supp:baseline}.

During the HitL deployment, a human operator controls the robot via keyboard inputs during high-uncertainty states.
Specifically, the operator provides a delta pose and gripper control, with each assistance event allowing control for $4$ steps. 
The results presented in this paper reflect human assistance commands rather than the number of expert calls, although these can be easily converted to the amount of expert calls if needed.
\begin{figure}[H]
    \centering
    \includegraphics[width=0.8\linewidth]{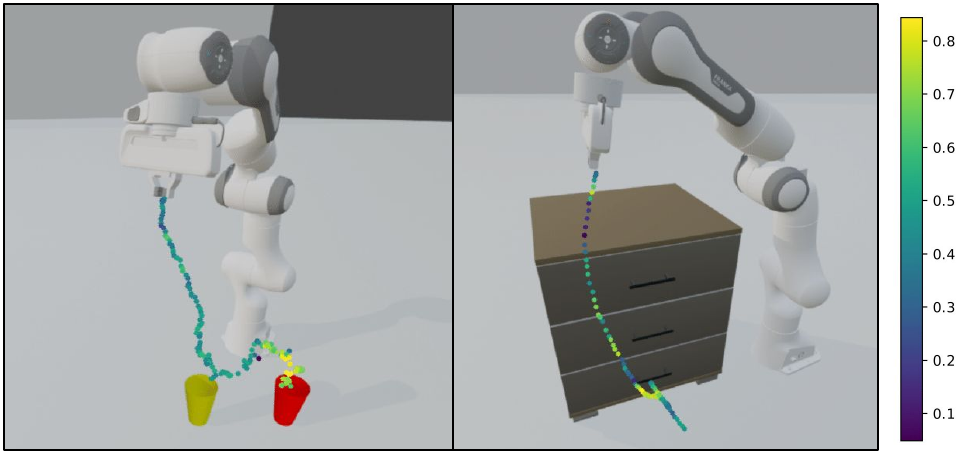}
    \caption{Qualitative visualization of predicted uncertainty, with lighter colors indicating higher uncertainty \label{fig:sim-qualitative}}
\end{figure}

\subsection{Simulated Environments}
\vspace{-0.3em}
We now summarize the simulated environments used to test our method; additional details about collected training data can be found in Table~\ref{tab:training_details}. \textbf{Distribution shift: Lift-sim.} In this task, we ask the robot to grasp and lift objects in a table-top setting. To emulate distribution shift, demonstration data is collected using only a single object (red cube - see Fig.\ref{fig:envs}), while for testing we roll out the pretrained policy to a set of unseen objects (round nuts, hammers, and hooks).
\textbf{Partial observability: Cup Stacking.} Here, we ask the robot to grasp a green cup and place it inside a red cup. We use three views as our observation: front, side, and wrist. Successful execution requires the robot to infer object alignment based on its observations. Misalignment can lead to unintended collisions, leading to failures. To introduce variability, cup positions are randomized during data collection. 
\textbf{Incorrect choice of action mode: Open drawer.} Here, the robot is tasked with opening one of three drawers in the scene. The collected dataset includes trajectories for opening each drawer, with $1/3$ of the data corresponding to each drawer. However, the dataset does not specify which drawer is to be opened in a given trajectory, introducing under-specification. 

As a sanity check, we first evaluate the unassisted task performance of the diffusion policy on each task under the training data distribution.
For the Lift-sim task, the fully autonomous policy achieves $100\%$ success rate on the training object but fails completely ($0\%$ success rate) on unseen objects. 
For Cup Stacking, the robot consistently picks up the first cup (100\% grasp success rate) but fails to place it into the second cup due to alignment difficulties, resulting in a success rate of $0\%$ without human assistance. This task is also sensitive to observation selection -- training with only side and front views causes the robot to fail when grasping the green cup.
For Open Drawer, the fully autonomous policy learns to open a drawer with $100\%$ success if the task description does not specify which particular drawer should be opened. Interestingly, despite the under-specified training (i.e., no conditioning on which drawer to open), the policy captures the multi-modality of the training distribution. During 100 rollouts with random sampling, the robot opens the middle and bottom drawers in 15\% and 85\% of trials, respectively, but never opens the top drawer.

\subsection{Efficiency of Human Interactions}

\begin{table}[t]
    \centering
    \setlength{\tabcolsep}{0.8pt}
    \caption{Average \# of human assistance steps needed to achieve 100\% success rates for simulated tasks. \label{tab:human-steps-sim}}
    \begin{tabular}{c|c c c}
        & \makecell{\small{Lift-}\\ \small{sim}} &\makecell{\small{Cup-}\\ \small{stacking}}& \makecell{\small{Open-}\\ \small{drawer}} \\ 
        \hline
        \small{\hula~\cite{he2024hula}} & \small{55.7($\pm$6.1)} & \small{54.0($\pm$16.3)} & \small{21.7($\pm$11.7)} \\
        \hline
        \small{\thriftydagger~\cite{hoque2021thriftydagger}} & \small{33.5($\pm$7.4)} & \small{21.2($\pm$15.6)} & \small{17.2($\pm$8.9)} \\
        \hline
        \small{\diffdagger~\cite{lee2024diff}} & \small{30.2($\pm$1.3)} & \small{32.0($\pm$4.0)} &  \small{16.0($\pm$4.4)} \\
        \hline
        \small{\ours} & \small{\textbf{16.9($\pm$4.5)}} & \small{\textbf{5.4($\pm$1.0)}} & \small{\textbf{8.0$(\pm$1.9)}}\\
        \hline
        \small{\fulltraj} &  \small{76.6($\pm$5.9)} & \small{147.8($\pm$12.9)} & \small{114.8($\pm$5.7)} \\
    \end{tabular}
    \vspace{-1.5em}
\end{table}

We now evaluate HitL deployment performance of these tasks. We note again that 100\% success rate is always possible with sufficient human assistance. Thus, we focus here on achieving high success rates with as few human assistance steps as possible, which is a critical aspect for real-world scalability of HitL systems. 

As shown in Table~\ref{tab:human-steps-sim}, for all simulated tasks, our method outperforms all baselines, and allows the policy to achieve perfect task success with the fewest intervention steps. Qualitatively (see accompanying video), we observe that, for Lift-sim, the robot only seeks human assistance when its gripper is close to the object, and lifting happens without intervention. For Cup Stacking, our method identifies states where the agent aligns the two cups as having high uncertainty, whereas picking up a cup (which benefits from unoccluded view) is marked as low uncertainty. Finally, for Open Drawer, the policy asks for assistance when it needs to decide which drawer to reach to, and, once the human operator steers it towards the intended target, the robot autonomously completes the rest of the task.

Looking at baselines, we find that \thriftydagger~provides good uncertainty estimation for in-distribution data (e.g. high state novelty when close to the first cup and when placing on the second cup for Cup Stacking), but its autonomous behavior is less effective and thus requires a low threshold for human assistance, leading to more interventions for a 100\% success rate. \diffdagger, which, like us, relies on diffusion models as a policy class, also requires more human assistance to achieve 100\% success rates. Finally, \hula~performs the worst, likely due to its inability to utilize a sparse reward in an offline setting.

Several key hyperparameters influence the performance of our system. We focus here on the most critical one -- the uncertainty threshold -- and defer discussion of others to Appendix~\ref{supp:hyperparams}. In this work, we always use an uncertainty metric threshold set at the 95$\%$ quantile of a held-out validation set not used in training or testing. We find that this selection consistently leads to 100$\%$ success rate with low teleoperator involvement. Figure~\ref{fig:finetune-lift} shows the effect of further lowering this threshold for the cup stacking task, which, as expected, leads to more human assistance. We note that all points shown in this figure represents 100$\%$ success rate over five rollouts (cup location randomized), which highlights that our metric can consistently detects critical states that requires human assistance.

\subsection{Fine-tuning Performance}
\begin{figure}
    \begin{center}
        \includegraphics[width=\linewidth]{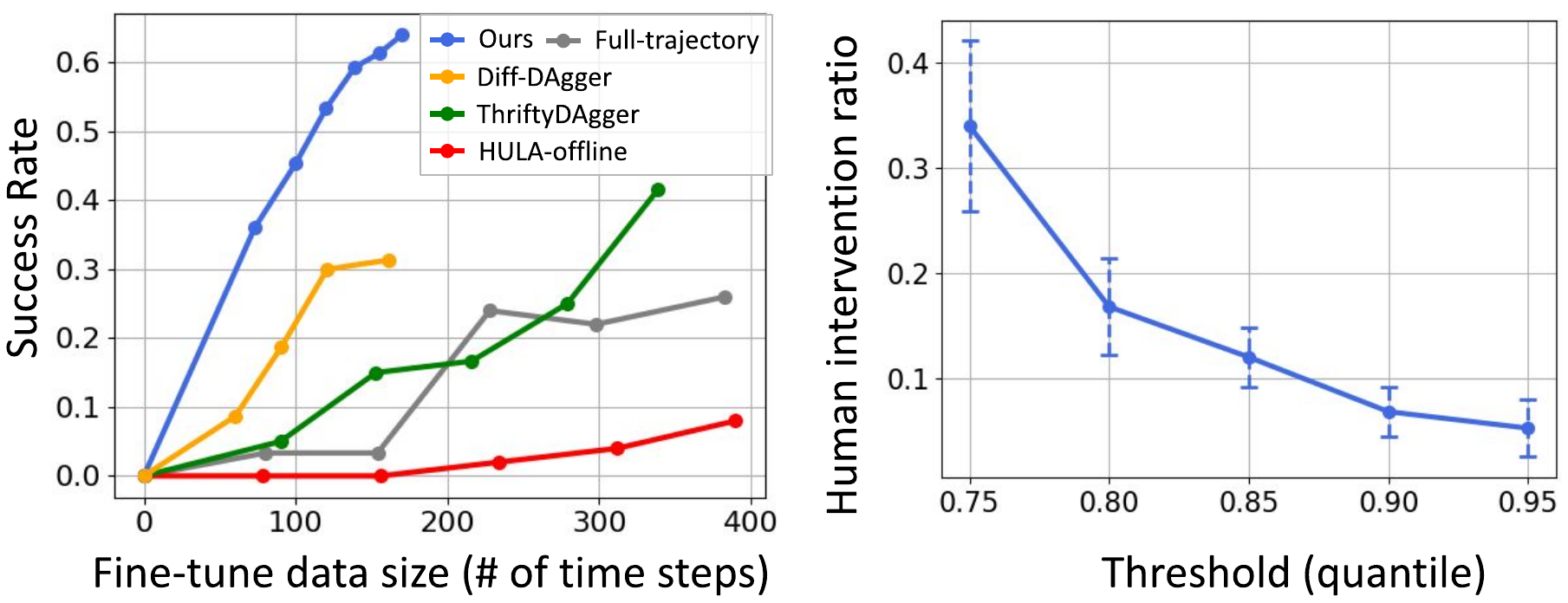}
    \end{center}
    \caption{Left: Average success rate of fine-tuning the Lift-sim task with different number of human intervention steps. Right: Sensitivity to threshold selection.\label{fig:finetune-lift} \vspace{-1em}}
\end{figure}

Our method requests operator assistance in states where the policy exhibits high uncertainty. We posit that these states are particularly valuable as they highlight areas where the policy can benefit from additional data collection for fine-tuning. We test this hypothesis by checking if leveraging our uncertainty metrics reduces the amount of data required for fine-tuning, while still achieving significant performance improvements.

Figure~\ref{fig:finetune-lift} shows autonomous policy performance improvements as a function of the size of the fine-tuning dataset, for our method as well as the baselines. Our method also consistently achieves higher success rates with the similar amount of fine-tuning data. Among all baselines, \diffdagger~shows best improvement with small amount of data. We explain this by its use of diffusion-based policies. We also note that fine-tuning on carefully curated data also outperforms the simple baseline of fine-tuning using full trajectories (i.e. complete additional demonstrations on testing scenarios). We note that the data used for each fine-tuning experiment is collected independently (i.e. the HitL fine-tuning dataset is not a part of the full-trajectory data set). For the HitL fine-tuning, the fine-tuning dataset only consists of actions when the robot is operated by human operators, instead of full trajectories.

\begin{figure}[ht!]
    \centering
    \includegraphics[width=0.85\linewidth]{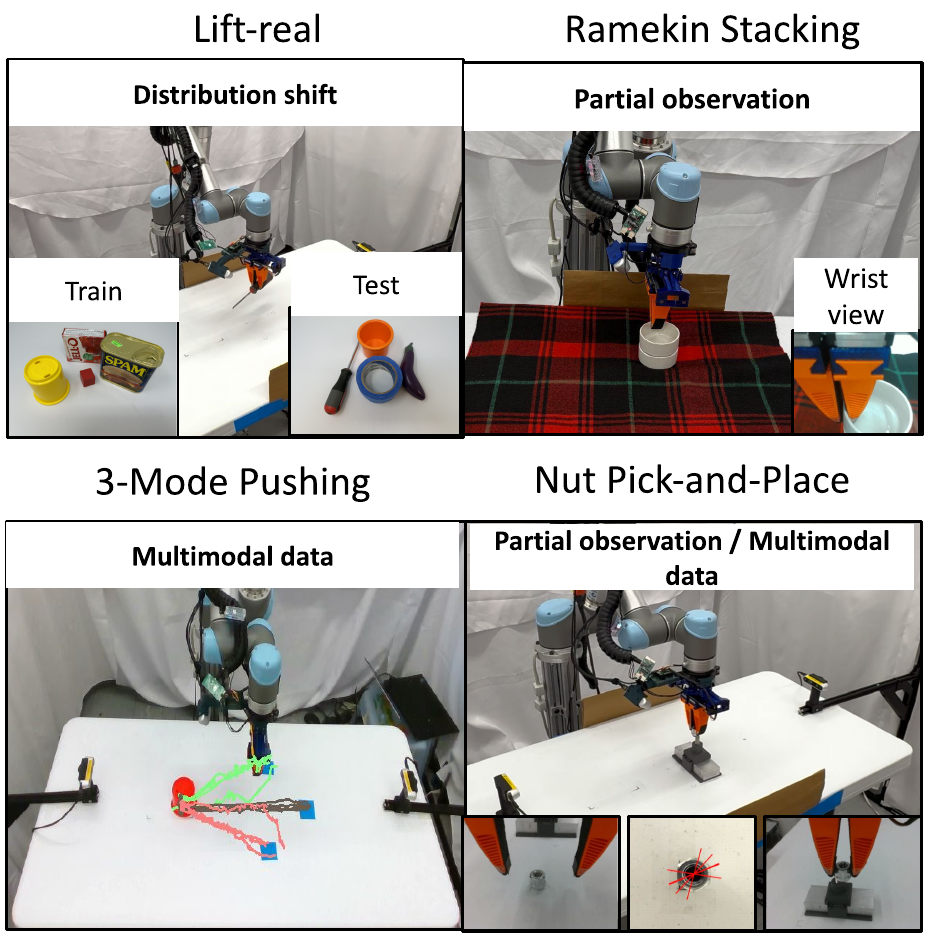}
    \caption{\textbf{Real robot experiments:} we design our experiments to elicit the challenges described in Sec.~\ref{sec:sim-exp} on a real robot.\label{fig:real-tasks}}
\end{figure}
\subsection{Real robot experiments}

\label{sec:real-exp}

Finally, we validate our method on real robot data collected via tele-operation. 
To support real-world deployment, we employ denoising diffusion implicit models (DDIM)~\cite{song2020denoising} for high-frequency action generation. 
We evaluate our method on 4 real robot tasks (see Fig.~\ref{fig:real-tasks}). As in the simulated experiments, we show an example in each of the deployment problems. We employ a transformer-based diffusion policy (DP-T) \cite{chi2023diffusion}, which yields stable performance across the tasks tested. 
For each experiment, we use the best-performing policy from pretraining since policies with lower performance typically require more human assistance.

\begin{table}[t]
    \centering
    \caption{Average \# of human assisted steps needed to achieve 100\% success rate during policy deployment.}
    \setlength{\tabcolsep}{0.7pt}
    \begin{tabular}{c| c c c c}
        & \small{Lift-real}& \small{Stacking} & \small{3-Mode Pushing} & \small{Nut PnP} \\ 
        \hline
        Ours & \small{7.2} & \small{6.8} & \small{6.5} & \small{8.4}\\
        \hline
        \\[-3mm]
       Full traj.& \small{80.0} & \small{111.9} & \small{98.9} & \small{48.8}
    \end{tabular}
    \label{tab:human-steps-real}
\end{table}

With HitL deployment, the robot can complete all four tasks. On average, our method only requests help from the human for approximately 8.3\% of time steps for an 100\% success rate (see Table~\ref{tab:human-steps-real}). We note that, since we are using action chunking during real robot deployment, one human intervention allows the human to control the robot for four steps, the same as the diffusion policy. Details about the evaluations protocol are included in Table~\ref{tab:training_details}.

\begin{table*}[t]
\centering
\caption{Training details: S, W, F represents side, wrist, front camera views receptively. L represents low-dimensional observations that contains end-effector pose and gripper state.\label{tab:training_details}}
\begin{tabular}{c|c|c|c|c|c|c|c}
 & \multicolumn{3}{c|}{Simulated tasks} & \multicolumn{4}{c}{Real tasks} \\ \hline
 & \makecell{Lift-\\sim} & \makecell{Cup\\stacking} &  \makecell{Open\\drawer} & \makecell{Lift-\\real} & \makecell{Ramekin\\stacking} & \makecell{3-mode\\pushing} & \makecell{Nut\\PnP} \\ 
 \hline
 Pretrain dataset size & 9666 & 41670 & 33617 & 9600 & 3360 & 5940 & 2450 \\
 \hline
 Observations & WFL & SWFL & FL & \multicolumn{4}{c}{\makecell{Two side cameras views,\\one wrist view and low-dim obs}} \\
 \hline
 Action chunk Size & \multicolumn{3}{c|}{1}  & \multicolumn{4}{c}{8} \\
 \hline
 Obs. history size & \multicolumn{3}{c|}{1}  & \multicolumn{4}{c}{2} \\
 \hline
 Image crop size & 76 x 76 & \multicolumn{2}{c|}{96 x 96} & \multicolumn{4}{c}{216 x 288} \\
 \hline
 Initial object position & \multicolumn{2}{c|}{randomized} & fixed & \multicolumn{2}{c|}{randomize}  & \multicolumn{2}{c}{fixed} \\
 \hline
 $\#$ of eval rollouts & \makecell{20 $\times$ 3} & 20 & 15 & \makecell{20 $\times$ 4} & 20 & 15 & 20
  \end{tabular}

\end{table*}
Qualitatively, our method identifies crucial states during policy execution. For example, in the Lift-real task, the robot asks for assistance when the gripper is close the the object. Using human-collected data with uncertainty, we can fine-tune the diffusion policy to improve 47\% success rate on average (shown in Table~\ref{tab:lift-real-finetune}), outperforming fine-tuning with full-trajectories of data. In the 3-Mode Pushing task, the robot autonomously reaches to the side of the object and then transfers control to the human operator, who poses the gripper in the correct location depending on the intended target. 
\begin{table}[H]
    \centering
    \caption{Fine-tuning performance of the Lift-real task. Results are success rates derived by 20 policy rollouts per object.}
    \label{tab:lift-real-finetune}
    \begin{tabular}{c|c c c}
         & \small{Train} & \small{Test} & \small{$||\mathcal{D}_{ft}||$} \\ 
        \hline
        \small{Zero-shot}  & \small{1} & \small{0.16} & \small{0} \\
        \hline
        \small{HitL fine-tuning (Ours)} & \small{1} & \small{0.63} & \small{80} \\
        \hline
        \small{Full-traj. fine-tuning} & \small{1} & \small{0.31}  & \small{132}  \\
    \end{tabular}
\end{table}

\begin{figure}[h]
    \centering
    \includegraphics[width=0.9\linewidth]{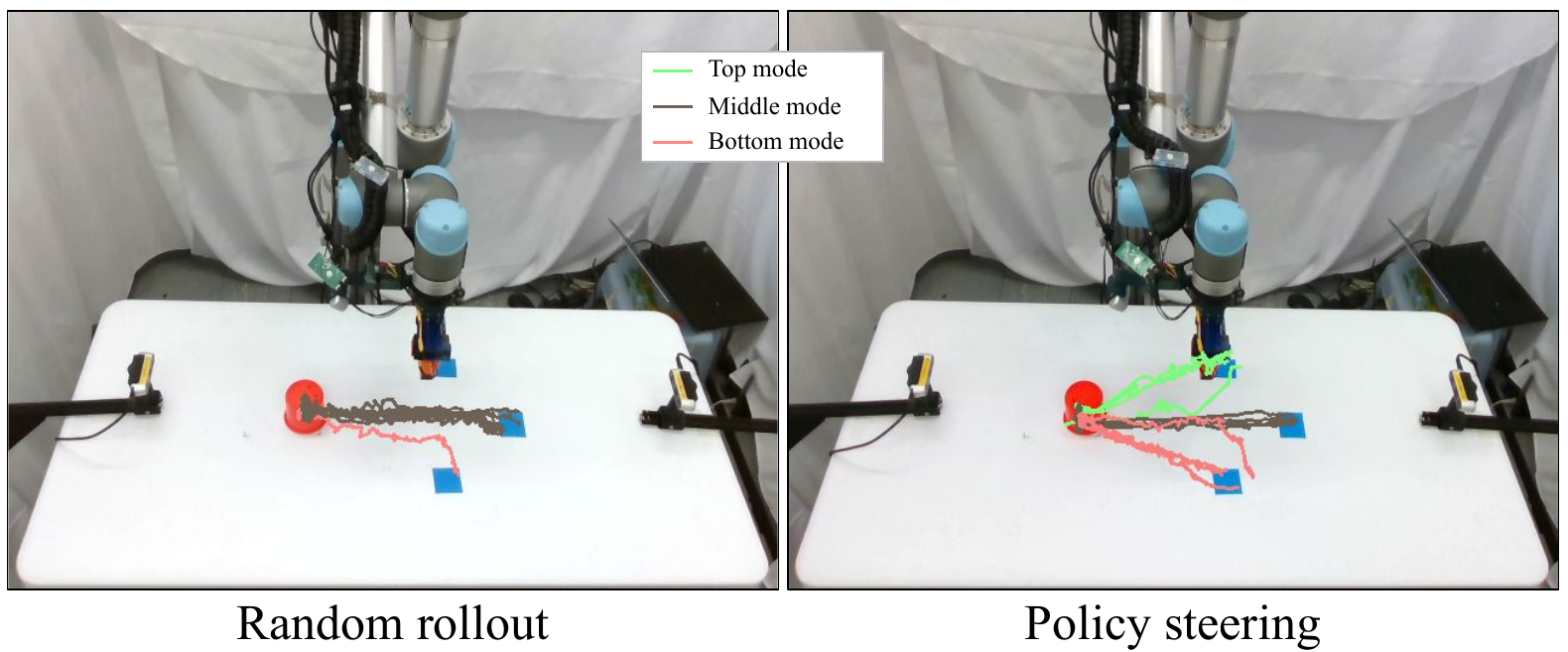}
    \caption{Object trajectories visualization: policy random rollout vs policy steering.}
    \label{fig:push-rollout-compare}
    \vspace{-1em}
\end{figure}
Once the pose of the gripper is indicative of the desired target, uncertainty drops, and the robot takes over and completes the task autonomously. In the Ramekin Stacking task, our method identifies high-uncertainty alignment states when the bottom ramekin is visually occluded (see Fig.~\ref{fig:real-tasks}). In contrast, grasping the first ramekin -- where visual observations suffice -- is marked as low-uncertainty and thus performed autonomously.

Finally, in the Nut Pick-and-Place task, our method assigns high uncertainty to two critical stages of execution: positioning for grasping (where the dataset contains diverse strategies for aligning the gripper with the nut edge as shown in Fig.~\ref{fig:real-tasks}) and placement (where precise positioning of the nut is required). The visual observations from the wrist camera and the two side cameras fail to reliably determine the stability of the placement, resulting in elevated uncertainty. The agent thus requests operator assistance for task completion.

\section{Hyperparameter exploration}
\label{supp:hyperparams}

In this section, we investigate how hyper-parameters affect the performance of our HitL agent.

\mystep{Sampling Radius.} The radius parameter $r$ defines the neighborhood size for collecting denoising vectors in uncertainty estimation. It is crucial to consider the scale of $r$ relative to the action distribution that a diffusion policy aims to recover. 
If $r$ is too large, the uncertainty metric becomes similar across all states, as it incorporates a broad range of them. Conversely, if $r$ is too small, the metric may be overly influenced by local uncertainty. In this work, we normalize each dimension of the action space to $[0,1]$, simplifying the selection of $r$.

In HitL fine-tuning for Lift-sim (see Table~\ref{tab:ablation-radius}), we test radii ranging from 0.01 to 0.1. 
A radius of 0.05 achieves the best balance between intervention steps and success rate, requiring only 20 interventions while improving the success rate by 0.63. This efficiency results from accurate uncertainty detection when the gripper approaches unseen targets but fails to grasp them. Smaller radii  and larger radii yield less precise estimations, leading to interventions that are either premature or delayed, reducing success rates despite more interventions.
Overall, a radius of 0.05 consistently achieves optimal performance by balancing local and global uncertainty estimations. Smaller radii miss key trajectory patterns, while larger radii incorporate irrelevant vectors, reducing the accuracy of uncertainty estimation in robotic manipulation tasks.

\mystep{Scaling constant} $\alpha$ in uncertainty calculation. The alpha parameter serves as a scaling factor in our uncertainty calculation, balancing two components: mode divergence and overall variance. Mode divergence captures directional differences between action modes, while overall variance measures spread within each mode.

Directional differences typically provide stronger uncertainty signals, and small $\alpha$ values (0.01 -- 0.1) emphasize mode divergence, effectively identifying critical occlusion phases (e.g., step 139). 
This highlights mode divergence as a reliable indicator of uncertain states. 
As $\alpha$ increases, the overall variance term gains more weight, raising both maximum and minimum variance values. 
However, this added emphasis on within-mode spread does not significantly enhance uncertain state detection, supporting the dominance of mode divergence as the more informative component.
The variance term remains essential for distinguishing states with single action mode (e.g. reaching for grasping), where mode divergence alone would yield identical scores. 
While $\alpha$ influences absolute uncertainty values, it minimally affects the identification of critical steps (consistently around steps 137–139 in cup stacking). 

\begin{table}
    \centering
    \caption{Effect of Sampling Radius on Fine-tuning Performance of the Lift-sim task.}
    \begin{tabular}{c|c c c c}
        Radius of sampling & 0.01 & 0.03 & 0.05 & 0.1\\ 
        \hline
        $\#$ of human steps ($\downarrow$)  & 60.3 & 31.6 & 20 & 46.3\\
        \hline
        Success rate ($\uparrow$) & 0.46 & 0.55 & 0.63 & 0.53\\
    \end{tabular}
    \label{tab:ablation-radius}
\end{table}

\begin{table}
    \centering
    \setlength{\tabcolsep}{2.5pt}
    \caption{Effect of scaling constant $\alpha$ on Fine-tuning Performance of the Lift-sim task.}
    \begin{tabular}{c|c c c c c}
        $\alpha$ & 0.01 & 0.05 & 0.1 & 0.3 & 0.5\\ 
        \hline
        $\#$ of human steps & 5 & 5 & 5 & 7  & 7\\
        \hline
        index of max uncertainty  & 139 & 139 & 139 & 137 & 137\\
        \vspace{-2em}
    \end{tabular}
    \label{tab:ablation-radius}
\end{table}

\section{Conclusions} 
\label{sec:conclusion}
We propose a method that enables robots to actively request HitL assistance during deployment using an uncertainty metric derived from diffusion policy denoising. This allows robots to identify states where human input is most valuable, reducing unnecessary monitoring while improving autonomy and reliability. Experiments show gains in policy performance, adaptability, and targeted data collection.
Future work will extend this approach to VLA models with diffusion-based action heads~\cite{black2024pi_0, bjorck2025gr00t} and explore interpretable feedback mechanisms that let robots convey uncertainty and intent intuitively, potentially leveraging Vision-Language Models~\cite{agia2024unpacking}. These directions aim to further automate interaction and scale HitL deployment in real-world robotics.

\bibliographystyle{IEEEtran}
\bibliography{references}
\label{sec:appendix}
\appendix
\subsection{Extra baselines}

\label{supp:baseline}
In this section, we also explore alternative methods that could potentially be adapted into a HitL pipeline. While promising in their original use cases, we found these methods unsuitable for online HitL settings. Therefore, we present our findings here rather than including them as baselines in our standardized comparisons. All evaluations are conducted on the Cup Stacking task.

\begin{figure}[h]
    \centering
    \includegraphics[width=\linewidth]{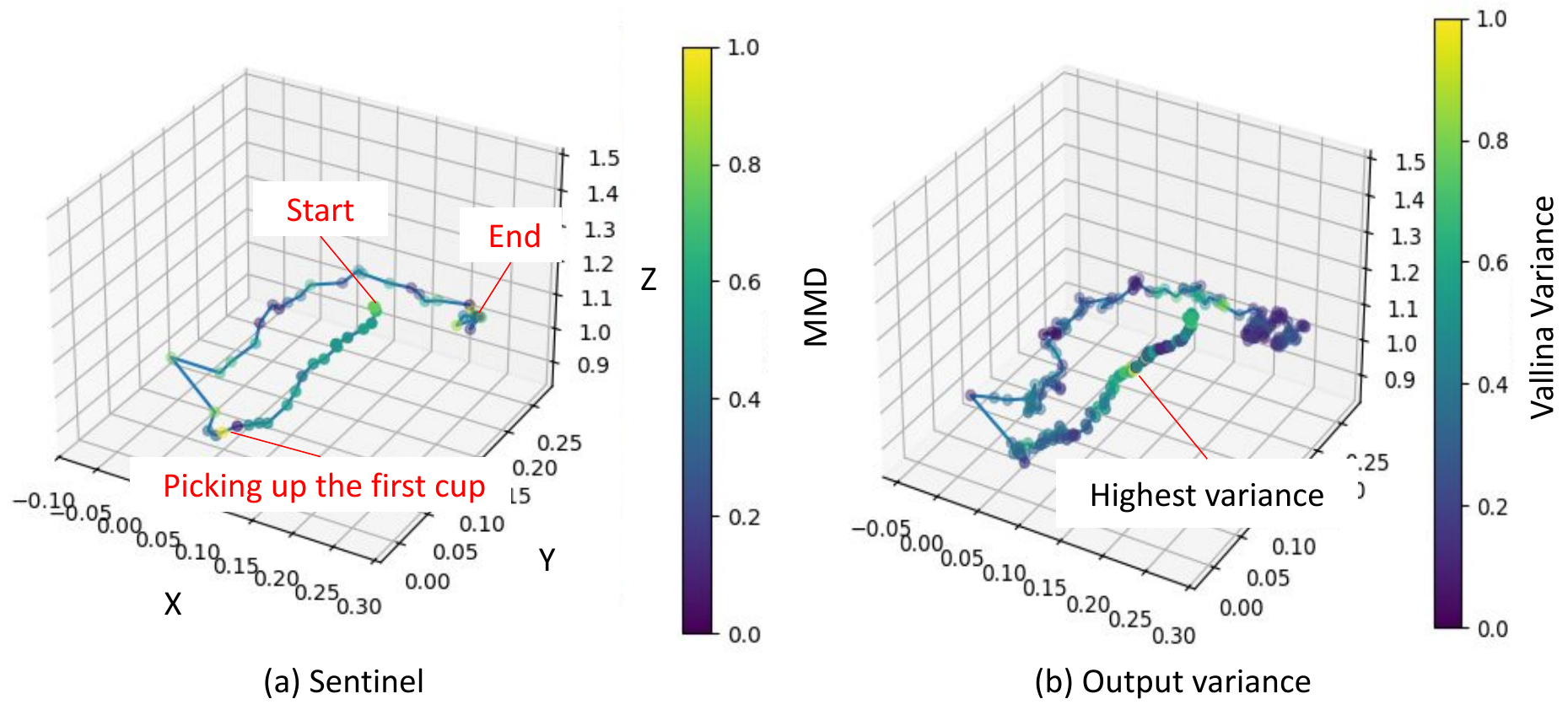}
    \caption{Qualitative results using (a) Sentinel~\cite{agia2024unpacking} and (b) output variance as metrics for HitL policy deployment.}
    \label{fig:stac-variance}
    \vspace{-1em}
\end{figure}

\mystep{Sentinel}~\cite{agia2024unpacking} is a method that utilizes action temporal consistency to analyze whether the agent is stopping to make progress. To compare with Sentinel, we train our diffusion policies with a longer action horizon of 8 steps, and use their Maximum Mean Discrepancy (MMD) with radial basis function (RBF) kernels to calculate in consistency. To derive best performance of Sentinel, we do a hyperparameter search on how many steps we should use to calculate MMD and select the best results based on whether it assigns the states that are visually occluded high MMD.

We observe that Sentinel performs poorly on our tasks for two main reasons. First, it often assigns high inconsistency to states involving reversed motions (e.g., approaching and lifting an object), even when the robot can complete these actions autonomously. This is due to its reliance solely on action inconsistency, which is naturally high in such transitions. Second, its OOD detection suffers from higher latency because it depends on past actions for computing consistency. While this makes Sentinel effective for failure detection, it is ill-suited for HitL policies, where delayed intervention can result in unrecoverable failures (e.g., knocking over another cup in the Cup Stacking task).

\mystep{Output Variance.} We also experimented with directly using output variance as a metric for HitL deployment. However, this approach proved unstable and often produced false negatives. For instance, in the Cup Stacking task, the model assigns the highest variance to early trajectory steps (e.g., approaching the first cup), while showing low variance during critical failures (e.g., cup insertion). This occurs because early stages exhibit high single-mode variance, whereas during insertion, the model outputs fall into two distinct modes, reducing single-mode variance. These results underscore the importance of accounting for the multi-modality present in human demonstrations.

\addtolength{\textheight}{-12cm}   
\end{document}